\documentclass[review]{elsarticle}

\usepackage{amsmath}
\usepackage{multicol}
\usepackage{amssymb}
\usepackage{amsfonts}
\usepackage{float}
\usepackage{color}
\usepackage{algorithm}
\usepackage[noend]{algpseudocode}

\usepackage{graphicx}
\usepackage{hyperref}
\usepackage{enumitem}
\newlist{steps}{enumerate}{1}
\setlist[steps, 1]{label = Step \arabic*:}

\journal{Journal of \LaTeX\ Templates}

\begin{document}

\begin{frontmatter}

\title{A fully dense and globally consistent 3D map reconstruction approach for GI tract to enhance therapeutic relevance of the endoscopic capsule robot}
\tnotetext[mytitlenote]{Fully documented templates are available in the elsarticle package on \href{http://www.ctan.org/tex-archive/macros/latex/contrib/elsarticle}{CTAN}.}

\author[rvt,else]{Mehmet Turan}
\author[rvt]{Yusuf Yigit Pilavci}
\author[rvt,red]{Redhwan Jamiruddin}
\author[hs]{Helder Araujo}
\author[else]{Ender Konukoglu}
\author[rvt]{Metin Sitti}
\address[rvt]{Physical Intelligence Department, Max-Planck Institute, Stuttgart, Germany}
\address[else]{Computer Vision Laboratory, ETH Zentrum, Zurich, Switzerland}
\address[red]{Biomedical Engineering Department, Martin-Luther Univeristy Halle-Wittenberg, Halle, Germany}
\address[hs]{Robotics Institute, Coimbro University, Portugal}

%\address[els]{Physical Intelligence Department, Max-Planck Institute, Stuttgart, Germany}

%\author[turan@is.mpg.de,mysecondaryaddress]{Elsevier Inc}



\begin{abstract}
In the gastrointestinal (GI) tract endoscopy field, ingestible wireless capsule endoscopy is emerging as a novel, minimally invasive diagnostic technology for inspection of the GI tract and diagnosis of a wide range of diseases and pathologies.  Since the development of this technology, medical device companies and many research groups have made substantial progress in converting passive capsule endoscopes to robotic active capsule endoscopes with most of the functionality of current active flexible endoscopes.  However, robotic capsule endoscopy still has some challenges.  In particular, the use of such devices to generate a precise three-dimensional (3D) mapping of the entire inner organ remains an unsolved problem.  Such global 3D maps of inner organs would help doctors to detect the location and size of diseased areas more accurately and intuitively, thus permitting more reliable diagnoses. To our knowledge, this paper presents the first complete pipeline for a complete 3D visual map reconstruction of the stomach.  The proposed pipeline is modular and includes a preprocessing module, an image registration module, and a final shape-from-shading-based 3D reconstruction module; the 3D map is primarily generated by a combination of image stitching and shape-from-shading techniques, and is updated in a frame-by-frame iterative fashion via capsule motion inside the stomach.  A comprehensive quantitative analysis of the proposed 3D reconstruction method is performed using an esophagus gastro duodenoscopy simulator, three different endoscopic cameras, and a 3D optical scanner.
\end{abstract}

\begin{keyword}
\texttt{Endoscopic Capsule Robot} \sep 3D map reconstruction \sep Frame stitching
%\MSC[2010] 00-01\sep  99-00
\end{keyword}

\end{frontmatter}

%\linenumbers

\section{Introduction}
Many diseases necessitate access to the internal anatomy of the patient for diagnosis and treatment.  Since direct access to most anatomical regions of interest is traumatic, and sometimes impossible, endoscopic cameras have become a common method for viewing the anatomical structure.  In particular, capsule endoscopy has emerged as a promising new technology for minimally invasive diagnosis and treatment of gastrointestinal (GI) tract disease.  The low invasiveness and high potential of this technology has led to substantial investment in their development by both academic and industrial research groups, such that it may soon be feasible to produce a capsule endoscope with most of the functionality of current flexible endoscopes.\\\\
Although robotic capsule endoscopy has high potential, it continues to face challenges.  In particular, there is no broadly accepted method for generating a 3D map of the organ being investigated.  This problem is made more severe by the fact that such a map may require a precise localisation method for the endoscope, and such a method will itself require a map of the organ, a classic chicken-and-egg problem.  The repetitive texture, lack of distinctive features, and specular reflections characteristic of the GI tract exacerbate this difficulty, and the non-rigid deformities introduced by peristaltic motion further complicate reconstruction algorithms.  Finally, the small size of endoscope camera systems implies a number of limitations, such as restricted fields of view, low signal-to-noise ratio, and low frame rate, all of which degrade image quality.  These issues, to name a few, make accurate and precise reconstruction a difficult problem and can render navigation and control counterintuitive.\\\\
Despite these challenges, accurate and robust three-dimensional (3D) mapping of patient-specific anatomy remains a tantalising goal.  Such a map would provide doctors with a reliable measure of the size and location of a diseased area, thus allowing more intuitive and accurate diagnoses.  In addition, should next-generation medical devices be actively controlled, a map would dramatically improve a doctor’s control in diagnostic, prognostic, and biopsy-like operations.  As such, considerable energy has been devoted to adapting computer vision techniques to the problem of in vivo 3D reconstruction of tissue surface geometry.\\\\
Two primary approaches have been pursued as workarounds for the challenges mentioned previously.  First, tomographic intra-operative imaging modalities, such as ultrasound (US), intra-operative computed tomography (CT), and interventional magnetic resonance imaging (iMRI) have been investigated for capturing detailed information of patient-specific tissue geometry.  However, surgical and diagnostic operations pose significant technological challenges and costs for the use of such devices, due to the need to acquire a high signal-to-noise ratio (SNR) in real-time without impediment to the doctor.  Another proposal has been to equip endoscopes with alternative sensor systems in the hope of providing additional information; however, these alternative systems have other restrictions that limit their use within the body.\\\\
This paper proposes a complete pipeline for 3D visual map reconstruction using only RGB camera images, with no additional sensor information.  This pipeline is arranged in a modular form, and includes a preprocessing module for the enhancement of the image quality, an image-stitching module allowing for registration and fusion of images, and a shape-from-shading module for reconstruction of 3D structures.  To our knowledge, this is the first such integration of image stitching with shape-from-shading, and our method also proposes a novel method for removing the specular reflections often found in endoscopy images.  The entire pipeline is designed for endoscope-typical low-resolution images with the goal of obtaining a 3D reconstruction of the organ under observation.  Comprehensive analysis of the method is performed using an esophagus gastro-duodenoscopy simulator, three different endoscopic camera models, and a 3D optical scanner; this analysis validates the method’s ability of to create a global 3D map of the stomach that is updated in a frame-by-frame, iterative fashion by capsule motion inside the stomach.  In sum, then, our method proposes a substantial contribution towards a more general and extensive use of the information that capsule endoscopes may provide.
%Figure 001
\begin{figure}
\includegraphics[width=0.90\textwidth]{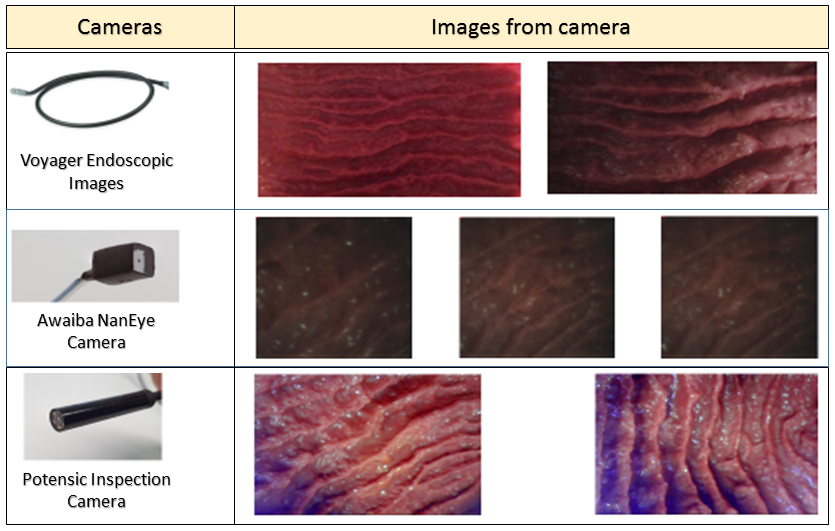}
\caption{Dataset overview for three different endoscopic cameras.}
\label{fig:01}       % Give a unique label
\end{figure}

\section{LITERATURE SURVEY}
Several previous studies have discussed 3D visual map reconstruction methods for standard and capsule endoscopes.  These methods may be broadly broken into for major types:
\begin{itemize}
\item	Stereoscopy, (stereo)
\item	Shape-from-shading (SfS)
\item	Structured light (SL)
\item	Time-of-flight (ToF)
\end{itemize}
Structured light and time-of-flight methods require additional sensors, with a concordant increase in cost and space; as such, they are not covered in this paper.
Stereo-based methods use the parallax observed when viewing a scene from two distinct viewpoints to obtain an estimate of the distance from the observer to the object under observation.  Typically, such algorithms have four stages in computing the disparity map \citep{10}: cost computation, cost aggregation, disparity computation and optimisation, and disparity refinement.\\\\
With multiple algorithms reported per year, computational stereo depth perception has become a saturated field. The first work reporting stereoscopic depth reconstruction in endoscopic images implemented a dense computational stereo algorithm \citep{11}.  Later, \citep{12} developed a semi-global optimization \citep{12}, which was used to register the depth map acquired during surgery to pre-operative models \citep{14}.  \citep{07} used local optimization to propagate disparity information around feature-matched seed points, \citep{16} and it has also been reported to perform well for endoscopic images \citep{17}.  This method was able to ignore highlights, occlusions or noise regions.\\\\
Despite the variety of algorithms and simplicity of implementation, computational stereo techniques bear several important flaws.  To begin, stereo reconstruction algorithms generally require two cameras, since the triangulation needs a known baseline between viewpoints.  Further, the accuracy of triangulation decreases with distance from the cameras due to the shrinkage of the relative baseline between camera centres and reconstructed points.  In endoscopy, these are important constraints; most endoscopes mount only one camera, and in those that mount more, the diameter of the endoscope inherently bounds the baseline.  As such, stereo techniques have yet to find wide application in endoscopy.\\\\
Due to the difficulty in obtaining stereo-compatible hardware, efforts have been made to adapt passive monocular three-dimensional reconstruction techniques to endoscopic images.  These techniques have been a focus of research in computer vision for decades, and have the distinct advantage of not requiring modification to existing endoscopic devices.  Two methods have emerged as useful in the field of endoscopy: Shape-from-Motion (SfM) and Shape-from-Shading (SfS).  Both methods have been demonstrated to have flaws: shape-from-shading often fails in the presence of uncertain information, e.g. bleeding, self-repetitiveness, and occlusions; shape-from-motion’s feature tracking methods tend to fail in the presence of repetitive tissue patterns.  Attempts to solve this latter problem with template-matching techniques have had some success, but tend to be too slow for real-time performance.\\\\
Shape-from-shading, which has been studied since the 1990s \citep{Horn} has demonstrated some suitability for endoscopic image reconstruction.  Its primary assumption is that the scene possesses a single light source, of which the intensity and pose relative to the camera is known, assumptions which are conveniently fulfilled in endoscopy \citep{19b11, 19b12, 19b13}. Further, the transfer function of the camera may be included in the algorithm in order to additionally refine estimates \citep{19b10}.  Additional assumptions are that the object reflects light in a Lambertian fashion and that the object surface has a constant albedo.  If these assumptions hold and the equation parameters are known, shape-from-shading can use the brightness of a pixel to estimate the angle between the camera’s depth axis and the shape normal at that pixel.  This has been demonstrated to be effective in recovering details, although global shape recovery often has flaws.\\\\
One additional barrier remains to 3D reconstruction in endoscopy, namely the visual complexity of scenes from endoscopic images.  Problems which are common in clinical images may cripple standard computer vision algorithms.  In particular, endoscopic algorithms must be robust to specular view-dependent highlights, noise, peristaltic movement, and focus-dependent changes in calibration parameters.  Unfortunately, a quantitative measure of algorithm robustness has not been suggested in literature, despite its clear value towards evaluation of algorithmic dependability and precision.\\\\
Our paper proposes a pipeline consisting of camera calibration, radial undistortion, reflection suppression, edge enhancement, de-vignetting, frame stitching, and shape-from-shading to reconstruct a 3D map of the organ under observation.  Amongst other contributions, an extensive quantitative analysis has been proposed and enacted to demonstrate the influence of each pipeline module on the accuracy and robustness of the reconstructed 3D map.  To our knowledge, this is the first such comprehensive mathematical and statistical analysis to be enacted in endoscopic image processing.

\section{METHOD}
\subsection{Dataset generation}
Our dataset was obtained on a non-rigid open GI tract model EGD (esophagus gastro duodenoscopy) surgical simulator LM-103 (cite our works here \citep{Meh, Meh1, Meh2}.  Paraffin oil was applied to the surface of the stomach model to imitate the mucosal layer in the stomach environment.  To ensure that our algorithm is not tuned to a specific camera model, three different endoscopic cameras were used for video capture.  The dataset was recorded in a controlled environment at the Max Planck Institute for Intelligent Systems.\\\\
We created a large dataset consisting of three different sub-datasets. A total of 5 hours of stomach video was recorded for this research, containing over a total of 9000 frames acquired by three cameras. The first sub-dataset, consisting of 3000 frames, was acquired with an AWAIBA NanEye camera (see Fig. 1 and Table 1) integrated into a robotic magnetically actuated soft capsule endoscope (MASCE) \citep{01, 21}      system; this system is actuated with electromagnetic coils. The second sub-dataset, consisting of 3000 frames, was acquired by integrating the POTENSIC inspection camera (see Fig. 1 and Table 2) of resolution 1280 x 720 pixels on our MASCE system with the specification shown in Table 2. Finally, the third sub-dataset, again of 3000 frames, was obtained by integrating the VOYAGER inspection camera (see Fig. 1 and Table 3) of resolution 720 x 480 pixels on our MASCE system with the specification shown in Table 3. We scanned the open stomach part of the simulator using the 3D Artec Space Spider image scanner and used this 3D scan as the ground truth for the error calculation for our 3D map reconstruction system (see Fig. 2).\\\\
In addition to these synthetic datasets, a capsule endoscope video of a patient’s stomach at UNSW Medical Department of Australia was provided.  This video was captured using the Olympus Endocapsule 10 capsule robot, and is 6 hours in length. This real dataset was used to test our method’s applicability to real endoscopic conditions after the quantitative analysis on the synthetic dataset.
%Table 01
\begin{table}
\caption{AWAIBA NANEYE MONOCULAR CAMERA SPECIFICATIONS}
\label{tab:1}       % Give a unique label
\begin{tabular}{ll}
\hline\noalign{\smallskip}
RESOLUTION & 250 x 250 Pixel \\
\hline\noalign{\smallskip}
FOOTPRINT & 2.2 x 1.0 x 1.7 MM \\
\hline\noalign{\smallskip}
PIXEL SIZE & 3 x 3 $\mu M^2$\\
\hline\noalign{\smallskip}
PIXEL DEPTH & 10 BIT\\
\hline\noalign{\smallskip}
FRAME RATE & 44 FPS\\
\hline\noalign{\smallskip}
\end{tabular}
\end{table}
%Table 02
\begin{table}
\caption{POTENSIC MINI MONOCULAR CAMERA SPECIFICATIONS}
\label{tab:1}       % Give a unique label
\begin{tabular}{ll}
\hline\noalign{\smallskip}
RESOLUTION & 1280 x 720 Pixel \\
\hline\noalign{\smallskip}
FOOTPRINT & 5.2 x 4.0 x 2.7 MM \\
\hline\noalign{\smallskip}
PIXEL SIZE & 10 x 10 $\mu M^2$\\
\hline\noalign{\smallskip}
PIXEL DEPTH & 10 BIT\\
\hline\noalign{\smallskip}
FRAME RATE & 15 FPS\\
\hline\noalign{\smallskip}
\end{tabular}
\end{table}
%Table 03
\begin{table}
\caption{VOYAGER MINI CAMERA SPECIFICATIONS}
\label{tab:1}       % Give a unique label
\begin{tabular}{ll}
\hline\noalign{\smallskip}
RESOLUTION & 720 x 480 Pixel \\
\hline\noalign{\smallskip}
FOOTPRINT & 5.2 x 5.0 x 2.7 MM \\
\hline\noalign{\smallskip}
PIXEL SIZE & 10 x 10 $\mu M^2$\\
\hline\noalign{\smallskip}
PIXEL DEPTH & 10 BIT\\
\hline\noalign{\smallskip}
FRAME RATE & 15 FPS\\
\hline\noalign{\smallskip}
\end{tabular}
\end{table}
%Figure 02
\begin{figure*}
\includegraphics[width=0.90\textwidth]{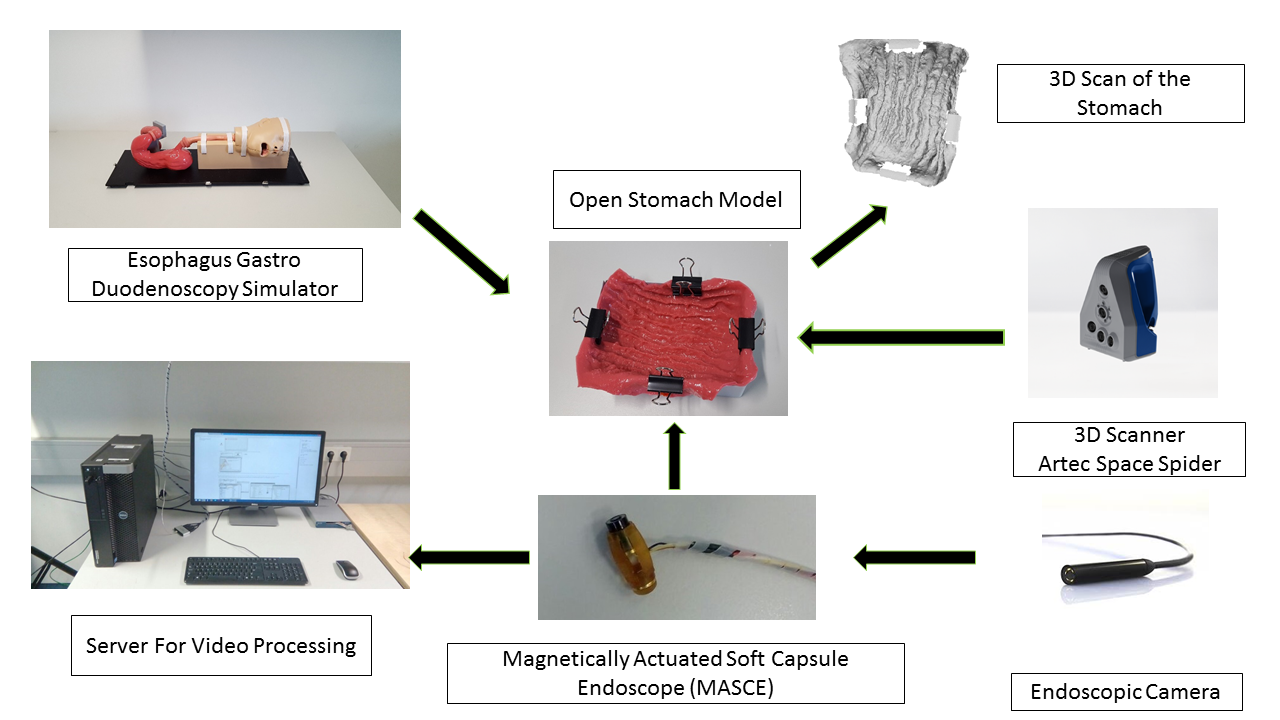}
\caption{Schematics of the experimental setup for 3D visual map reconstruction of an esophagus gastro duodenoscopy simulator for surgical training, open surgical stomach model, 3D image scanner, endoscopic camera, and active robotic capsule endoscope.}
\label{fig:02}       % Give a unique label
\end{figure*}

\subsection{Pre-processing}
The proposed 3D visual map reconstruction framework shown in Fig.3 and Fig.4 starts with a preprocessing module that suppresses reflections caused by inner organ fluids, and enhances image details to improve feature extraction and matching performance of the next module. We propose an original method for reflection detection and suppression, as illustrated in Fig.5. Eliminating specular reflections is a crucial preprocessing step due to the negative effect of such reflections on the performance of the image stitching procedure and shape-from-shading 3D reconstruction methods.
%Figure 03
\begin{figure*}
  \includegraphics[width=0.90\textwidth]{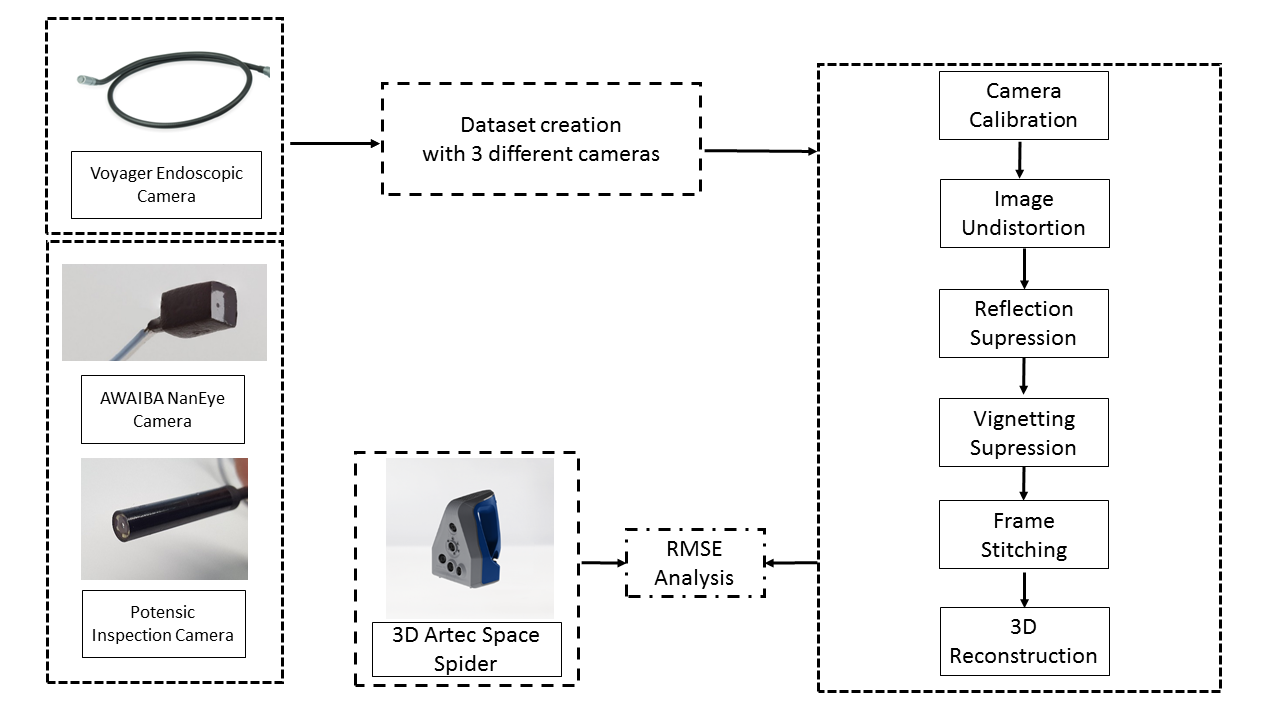}
\caption{The overview of the computational framework for 3D visual map reconstruction.}
\label{fig:03}       % Give a unique label
\end{figure*}
%Figure 04
%Figure 06
\begin{figure*}
  \includegraphics[width=0.90\textwidth]{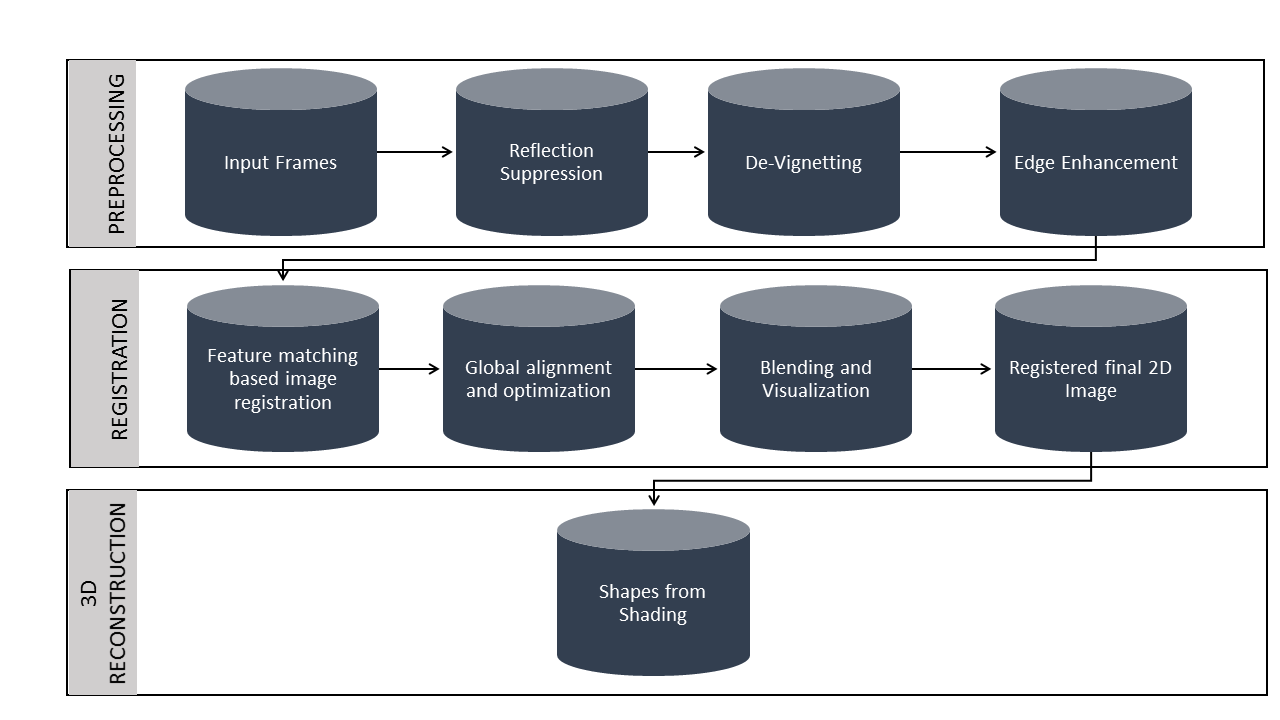}
\caption{ System Overview.}
\label{fig:06}       % Give a unique label
\end{figure*}

\begin{figure*}
  \includegraphics[width=0.90\textwidth]{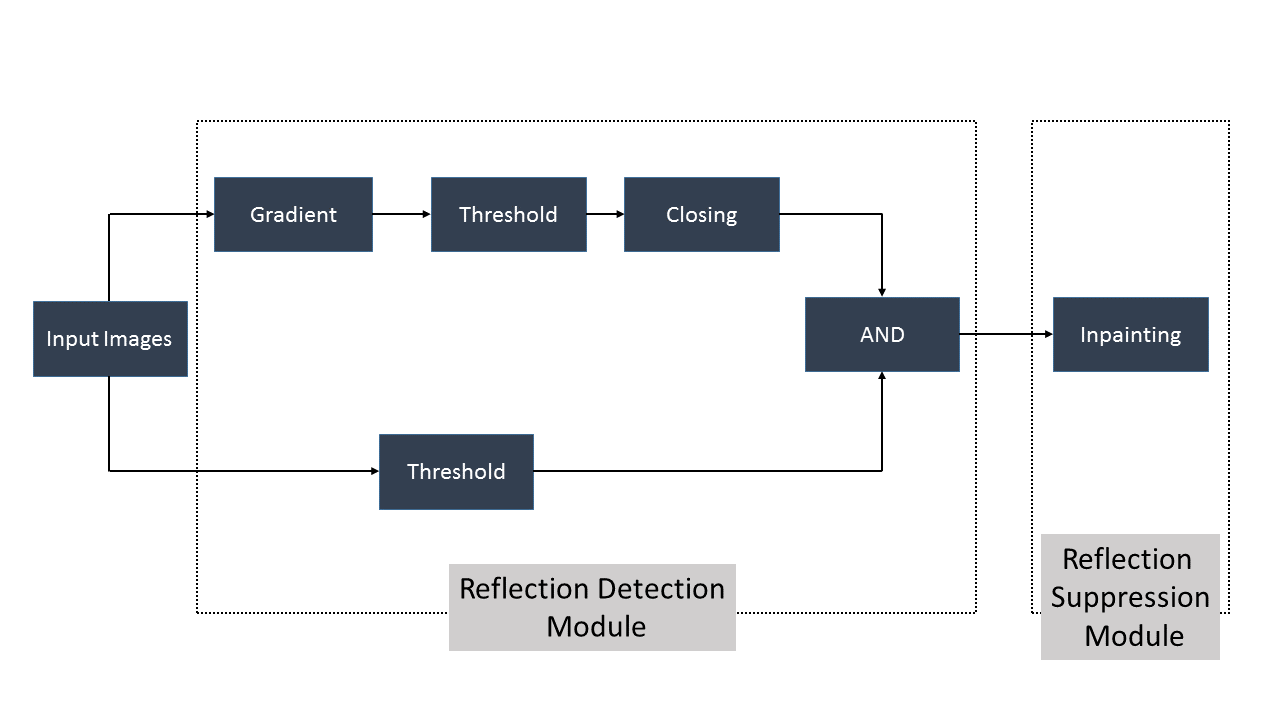}
\caption{Flowchart of the proposed light reflection detection and suppression method.}
\label{fig:04}       % Give a unique label
\end{figure*}
  We propose to detect specular reflections by combining the gradient map of the input image with the peak values detected by an adaptive threshold operation.  For this purpose, the magnitude of the image gradient is calculated and a threshold operation applied to the gradient map:
%Equation{1}
\begin{equation}
filter = \Bigg\{ 
\begin{tabular}{c c}
1 &  ,$M_i > Threshold$ \\
0 & ,otherwise \\
\end{tabular} 
\end{equation}
The filter defined by eq. (1) eliminates image regions with low gradient magnitudes by using an adaptive threshold.  A morphological filling operation is then applied to the closed regions to determine the areas affected by specularities. To guarantee closed regions, a morphological closing operation is applied.\\\\
Following this step an adaptive threshold method is applied by using the mean and standard deviation of the grey levels of the image:
%Equation{2}
\begin{equation}
Mask_\textrm{$Illumination_i$ = }\Bigg\{ 
\begin{tabular}{c c}
0 & ,$I_i < \mu_I+\sigma_I$\\
1 & ,otherwise \\
\end{tabular}  
\end{equation}
where $I_i$ is the grey level value of ith pixel in image $I_i$ and $\mu_I$, and $\sigma_I$ are the mean and standard deviation, respectively, of image $\textit{I}$. Combining the thresholded gradient map and this intensity-based threshold map locates pixels of specular reflections.\\\\
The next step after reflection detection is suppression of these reflection-distorted areas. For reflection suppression, the inpainting method is used. An overview of the method used to locate and suppress specularities is shown in Fig.5. 

\subsection{Camera calibration, correction of the lens distortions and vignetting cancellation}

A chessboard calibration method was applied to calculate intrinsic, extrinsic, and distortion parameters of the camera \citep{24}.  Extrinsic and intrinsic parameters are used to transform between 3D world coordinates and 2D image coordinates. Additionally, distortion parameters calculated by the camera calibration process are used to remove radial and tangential lens distortions.  Since endoscopic camera frames suffer from a high amount of radial distortions, estimation of these parameters can be quite important for the accuracy of the final map. After estimating distortion parameters, the Open CV function undistort was used to eliminate radial distortions from the images.\\\\
Vignetting is another important issue in endoscopic image processing.  Vignetting refers to the issue of inhomogeneous illumination distribution on the image corners with respect to image center, and is primarily caused by camera lens imperfections and light source limitation.  3D reconstruction and image stitching methods are generally very sensitive to such inhomogeneous illumination, so a robust vignetting correction is required before proceeding to those steps.\\\\
This paper applies a image vignetting correction based on the radial gradient symmetry as calculated via the image gradient from the center to the corners of the image \citep{22}.  The radial gradient value gives reliable information about the vignetting effect since the image brightness also shows a radially decreasing characteristic from image center to corners of the image. Radial gradient at point (x, y) in an image I can be computed by:
%equation{5}
\begin{equation}
\varphi_r^I (x,y) = \Bigg\{ 
\begin{tabular}{c c}
$\frac{|\bigtriangledown I(x,y).r(x,y)|}{|r(x,y)|}$ &, $|r(x,y)| >$ 0;\\
0 & ,$|r(x,y)|$ = 0 
\end{tabular}  
\end{equation}
%equation{6}
\begin{equation}
\bigtriangledown I(x,y) = 
[\frac{\partial I}{\partial x}, \frac{\partial I }{\partial y}]^T, r(x,y) = [x - x_0,y - y_0]
\end{equation}
Vignetting-corrected images display a symmetry in the radial gradient distribution, which is lacking in images with vignetting.  The method, then, corrects vignetting by enforcing the symmetry of the radial gradient. An example of a raw and a vignette-corrected image may be seen in Fig.6. After vignetting correction step, an unsharp masking filter is applied to enhance the details in the image and to sharpen it.
%Figure 05
\begin{figure*}
\begin{center}
  \includegraphics[width=0.5\textwidth]{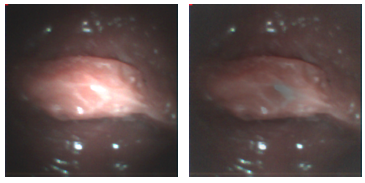}
\end{center}
\caption{The left image is the input image and the right one is the vignetting corrected image as we implemented the mentioned algorithm on a test frame.}
\label{fig:05}       % Give a unique label
\end{figure*}

\subsection{Frame stitching and Shape from Shading}
%Unlike the existing methods in literature, which create only partial, frame-by-frame 3D maps, our method reconstructs a global 3D map of the inner organ. In order to accomplish this, our method stitches the frames in the 2D spatial domain and uses the final stitched frame as the input for the 3D map reconstruction module (see Fig.9 and Fig.10).
Unlike the existing methods in literature, which create only partial, frame-by-frame 3D maps of the GI tract, our method reconstructs a globally consistent 3D map of the inner organ. In order to accomplish this, our method stitches the frames in the 2D spatial domain and uses the final stitched frame as the input for the 3D map reconstruction module (see Fig.8).

%Figure 07
\begin{figure*}
  \includegraphics[width= 0.90\textwidth]{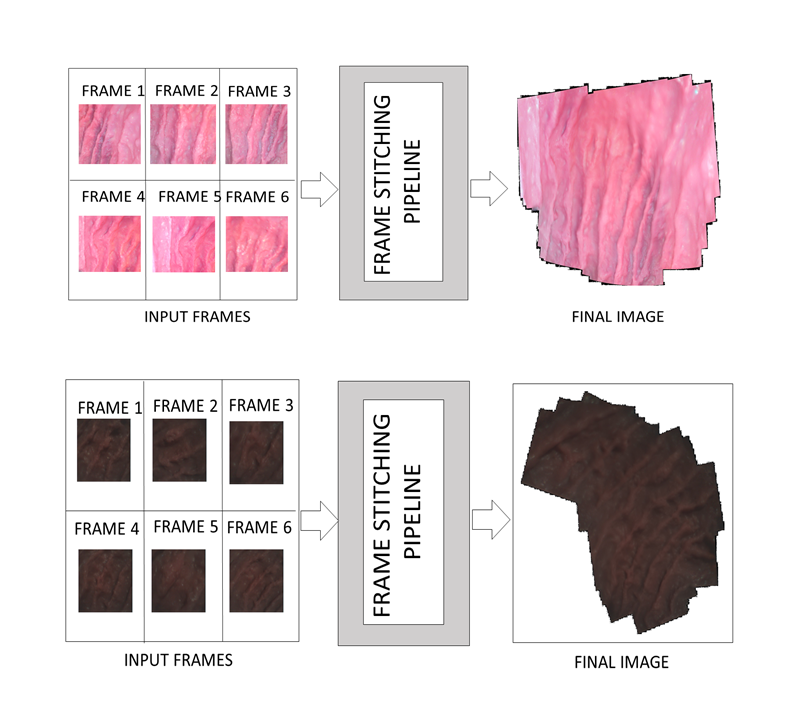}
\caption{ Demonstration of the frame stitching process for Potensic and NanEye camera frames.}
\label{fig:07}       % Give a unique label
\end{figure*}
%Figure 08
%\begin{figure*}
%% Use the relevant command to insert your figure file.
%% For example, with the graphicx package use
%  \includegraphics[width=0.90\textwidth]{figures/re1.png}
%\caption{Demonstration of the frame stitching process for Olympus capsule endoscope example 1.}
%\label{fig:08}       % Give a unique label
%\end{figure*}

An essential part of a stitcher module is the feature extraction step. To determine which feature descriptor is performing most accurately for endoscopic images, we evaluated the most modern and commonly used feature extraction and matching combinations i.e. SURF, SIFT ORB, HOG, MinEigenValues, dense SIFT and dense SURF on our endoscopic images dataset. We compared these feature extraction and description methods in respect of their matching capability on endoscopic images. For the mathematical evaluation of the matching capabilities, we calculated the reprojection error between matched points. Algorithm 1 shows the steps of the reprojection error calculation between mathced points.

\begin{algorithm}
\caption{Pseudo code to calculate the reprojection error between matched points}
\label{algo:calc_loss}
\begin{algorithmic}[1]
\State Extract and match feature points between two images using the selected feature descriptor.
\State Extract the locations of matched key points in both images.
\State Use these matched key point locations and the intrinsic camera matrix to find the perspective transformation matrix of this image pair.
\State Use the perspective transformation matrix to reproject the key point locations from the second image onto the first image.
\State Calculate the reprojection error between the reprojected and initial key point locations.
\end{algorithmic}
\end{algorithm}

The reprojection error analysis showed that dense SURF outperforms all other existing feature descriptors regarding accuracy of the matched feature points on endoscopic images. Another contribution of our paper is a novel frame stitching module inspired by the Open CV Stitcher class and developed considering the typical challenges faced by endoscopic image processing applications. Fig.4 illustrate the stitching module pipeline of our method. Unlike the stitcher class of Open CV, our method takes both the translation and rotation of the camera into account using the sparse bundle adjustment method. Algorithm 2 demonstrates the steps of our endoscopic stitching module. Fig.7 shows the stitching results.

\begin{algorithm}
\caption{Endoscopic stitching module}
\label{algo:calc_loss}
\begin{algorithmic}[1]
\State Use dense SURF to select the m candidate frames with the most feature matches with the current frame.
\State Use random sample consensus (RANSAC) to estimate image transformation parameters with a minimal set of randomly sampled correspondences and to find geometrically consistent feature matches.
\State Estimate camera parameters using the camera calibration information intrinsic matrix and feature matches.
\State Refine the parameters using sparse bundle adjustment method.
\State Perform multi-blending of frames and determine the connected components of each frame:
\begin{enumerate}
\item	For each connected component, apply bundle adjustment to solve for rotation and translation parameters.
\item	Re-estimate these parameters using the Levenberg-Marquardt method [(Elhabian 2008)].
\item  Finally, render the final stitched image using multi-band blending.
\end{enumerate}
\end{algorithmic}
\end{algorithm}

%Fig.7 and Fig.8 show the stitching results achieved for real and synthetic endoscopic videos.

%Figure 09
\begin{figure*}
\begin{center}
  \includegraphics[width=1\textwidth]{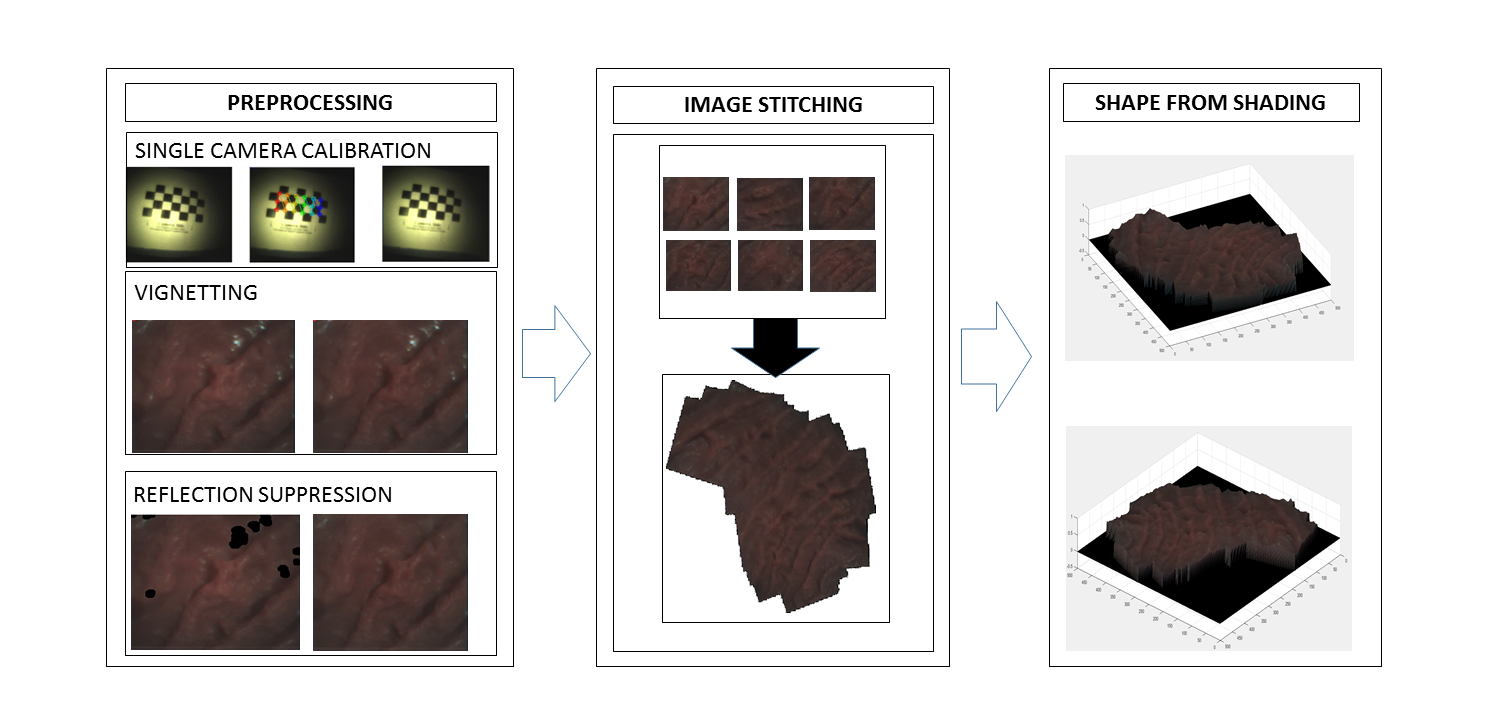}
\caption{ Demonstration of calibration, vignetting suppression, reflection detection, suppression, image stitching and shape from shading AWAIBA Camera Dataset.}
\label{fig:09}       % Give a unique label
\end{center}
\end{figure*}
%Figure 10
%\begin{figure*}
%% Use the relevant command to insert your figure file.
%% For example, with the graphicx package use
%  \includegraphics[width= 0.90\textwidth]{figures/fig14-Copy.png}
%\caption{ Demonstration of image stitching and shape from shading Olympus Capsule Endoscope dataset}
%\label{fig:10}       % Give a unique label
%\end{figure*}

After the estimation of the 2D stitched map, shape-from-shading is employed in order to create a 3D map of the entire organ at the millimetre scale.  In this paper, we employed the method of Tsai and Shah [(Ping-Sing and Shah 1994)], which is based on the following assumptions:
\begin{itemize}
\item	The object surface is lambertian
\item	The light comes from a single point light source
\item	The surface has no self-shaded areas. 
\end{itemize}
This first assumption is not obeyed by raw endoscopy images due to the specular reflections inside the organs.  We addressed this problem through the reflection suppression technique previously described.  This done, the above assumptions allow the image intensities to be modelled by
%Equation{8}
\begin{equation}
I(x,y)=\rho(x,y,z)*cos\Theta_i
\end{equation}
where $\textit{I}$ is the intensity value, p is the albedo (the reflecting power of surface), and $\theta_i$ is the angle between surface normal $\textit{N}$ and light source direction $\textit{S}$. With this equation, the grey values of an image I are related only to albedo and angle $\theta_i$.  Using these assumptions, the above equation can be rewritten as follows:
%Equation{9}
\begin{equation}
I(x,y)=\rho*N.S
\end{equation}
where (.) is the dot product, N is the unit normal vector of the surface, and S is the incidence direction of the source light.  These may be expressed respectively as
%Equation{10}
\begin{equation}
N=\frac{(-p(x,y),-q(x,y),1)}{(p^2+q^2+1)^\textrm{(1/2)}}
\end{equation}
%Equation{11}
\begin{equation}
S=(cos\tau*sin\sigma,sin\tau*sin\sigma,cos\sigma  )
\end{equation}
where ($\tau$) and ($\sigma$) are the slant and the tilt angles, respectively, and p and q are the x and y gradients of the surface Z:
%Equation{12}
\begin{equation}
p(x,y) = \frac{\partial Z(x,y)}{\partial x}
\end{equation}
%Equation{13}
\begin{equation}
q(x,y) = \frac{\partial Z(x,y)}{\partial y}
\end{equation}
The final function then takes the form
%Equation{14}
\begin{multline}
I(x,y) = \rho * \frac{(cos\sigma+p(x,y)*cos\tau*sin
\sigma+q(x,y)*sin\tau*sin\sigma  )}{((p(x,y))^2+((x,y))^2+1)^\textrm{(1/2)}} 
\\= R(p_\textrm{x,y}, q_\textrm{x,y})
\end{multline}
%Figure 11
%\begin{figure*}
%\begin{center}
%% Use the relevant command to insert your figure file.
%% For example, with the graphicx package use
%  \includegraphics[width=0.75\textwidth]{figures/fig12.png}
%\caption{ Alignment of reconstructed surface from 1 frames and ground truth.}
%\label{fig:11}       % Give a unique label
%\end{center}
%\end{figure*}
Solving this equation for t, p and q essentially corresponds to the general problem of shape from shading.  The approximations and solutions for p and q give the reconstructed surface map Z.  The necessary parameters are tilt, slant and albedo, and can be estimated as proposed in \citep{25}.  These parameters are necessary for the Tsai-Shah shape-from shading-approach.  The unknown parameters of the 3D reconstruction are the horizontal and vertical gradients of the surface Z, p and q. With discrete approximations, they can be written as follows:
%Figure12
%\begin{figure*}
%\begin{center}
%% Use the relevant command to insert your figure file.
%% For example, with the graphicx package use
%  \includegraphics[width=0.75\textwidth]{figures/fig13.png}
%\caption{ Alignment of the reconstructed surface from 10 frames and ground truth.}
%\label{fig:12}       % Give a unique label
%\end{center}
%\end{figure*}
%Figure11
%\begin{figure*}
%\begin{center}
%% Use the relevant command to insert your figure file.
%% For example, with the graphicx package use
%  \includegraphics[width=0.75\textwidth]{figures/fig14.png}
%\caption{ Alignment of reconstructed surface from 100 frames and ground truth.}
%\label{fig:11}       % Give a unique label
%\end{center}
%\end{figure*}
%Equation{15}
\begin{equation}
p(x,y)=Z(x,y)-Z(x-1,y)
\end{equation}
%Equation{16}
\begin{equation}
q(x,y)=Z(x,y)-Z(x,y-1)  
\end{equation}
where Z(x,y) is the depth value of each pixel.  From these approximations, the reflectance function R($p_\textrm{x,y}, q_\textrm{x,y}$) can be expressed as 
%Equation{17}
\begin{equation}
R(Z(x,y)-Z(x-1,y),Z(x,y)-Z(x,y-1))         
\end{equation}
Using equations [15], [16], and [17], the reflectance equation may also be written as
%Equation{18}
\begin{multline}
f(Z(x,y),Z(x,y-1),Z(x-1,y),I(x,y))\\
= I(x,y)- R(Z(x,y)-Z(x-1,y),Z(x,y)-Z(x,y-1))\\
=0
\end{multline}
Tsai and Shah proposed a linear approximation using a first-order Taylor series expansion for function f and for depth map $Z^\textrm{n-1}$, where $Z^\textrm{n-1}$ is the recovered depth map after n-1 iterations. The final equation is
%Equation{19}
\begin{equation}
Z^n  (x,y)=Z^\textrm{(n-1)} (x,y)  -\frac{f(Z^\textrm{(n-1)} (x,y))}{\frac{\mathrm{d}(f(Z^\textrm{(n-1)} (x,y))}{\mathrm{d}(Z(x,y))}}
\end{equation}
where f is a defined function, constrained by
%Equation{20}
\begin{equation}
\frac{\mathrm{d}f(Z^\textrm{(n-1)}(x,y))}{\mathrm{d}Z(x,y)}\sqrt{(1+i_x^2+i_y^2 )})
\end{equation}
and
%Eqaution{21}
\begin{equation}
i_x = cos\tau*\frac{sin\sigma}{cos\sigma}
\end{equation}
%Equation{22}
\begin{equation}
i_y = sin\tau*\frac{sin\sigma}{cos\sigma}
\end{equation}
%Equation 24, 25, 26
The nth depth map $Z^n$ is calculated by using the estimated slant, tilt, and albedo values. %\textcolor{Red}{Figure 12 shows the results of 3D map reconstruction for a group of stitched endoscopic images.}
%Figure 12
%\begin{figure*}
%% Use the relevant command to insert your figure file.
%% For example, with the graphicx package use
%  \includegraphics[width=0.75\textwidth]{figures/fig15.png}
%\caption{ 3D reconstructed stomach simulator surface.}
%\label{fig:12}       % Give a unique label
%\end{figure*}

\section{TESTING AND RESULTS}
As emphasized in the introduction, medical image 3D reconstruction papers in literature suffer from a lack of quantitative analysis.  We fill this gap in order to demonstrate the effect of a wide variety of common techniques such as image registration, shape-from-shading, and preprocessing techniques such as vignetting correction, edge enhancement and reflection suppression, on 3D map accuracy in a quantitative fashion.  In this section, we will discuss the 3D reconstruction precision of the proposed pipeline from different perspectives.\\\\
We used a 3D optical scan of the EGD Simulator acquired by Artec Space Spider as our ground truth for quantitative error calculations.  The Artec Space Spider is a 3D scanner with a resolution of up to 0.1 mm and a scan rate of 7.5 frames per second, ensuring precise and accurate information for the evaluation of the proposed framework. %\textcolor{Red}{Fig.11 illustrate the overlapping of the 3D reconstructed map for 100 frames over the 3D map of Artec Space Spider; this serves as the ground truth for the ICP method \citep{23}.} 
For the evaluation of the 3D reconstruction error, we created three types of groups; large groups consisting of 100 frames, medium groups consisting of 50 frames and small groups consisting of 1 frame, respectively. For all of the evaluation groups, the root mean square (RMS) error between the depth value $d_j$ of each pixel from the source point cloud and the depth value $d_\textrm{$\overline{j}$}$ for the corresponding pixel from the reference point cloud was calculated using:
%Equation{23}
\begin{equation}
  RMS error = \sqrt{\frac{1}{N}\sum_{j=1}^{N}(d_j - d_\textrm{$\overline{j}$})}  
\end{equation}
%changed fig 13 to fig 9
where $d_j$ and $d_\textrm{$\overline{j}$}$ are in millimeters.  The mean and standard deviation of the RMS errors were calculated using these RMS values. Fig. 9 shows the error rates for different groups and different camera types. In general, NanEye camera has the worst performance, followed by Voyager and Potensic.  The main reason for this performance variance is resolution and general image quality; higher resolution and image quality results in more accurate image stitching, less reflection, less vignetting artefacts, sharper images, and less noise. All of these factors heavily affect the accuracy of the 3D map reconstruction process.\\\\
Another important point is that error rate is directly proportional to group size: large group (100 frames), medium group (50 frames) and small group (1 frame), respectively.  This increase can be traced to the cumulative errors that the 2D pairwise frame-stitching process introduces; the stitching of each frame pair introduces a certain amount of error, and as the number of stitched frames increases, the error increases with it.  For large point clouds consisting of 100 frames, we achieved 5.1$\%$ mean reconstruction error with a standard deviation of 1.1$\%$. For smaller point clouds of 50 frames and 1 frame the means of the errors are 2.6$\%$ and 2.2$\%$ with standard deviations of 0.25$\%$ and 0.12$\%$, respectively.\\\\
%Figure 15
\begin{figure*}
  \includegraphics[width=1\textwidth]{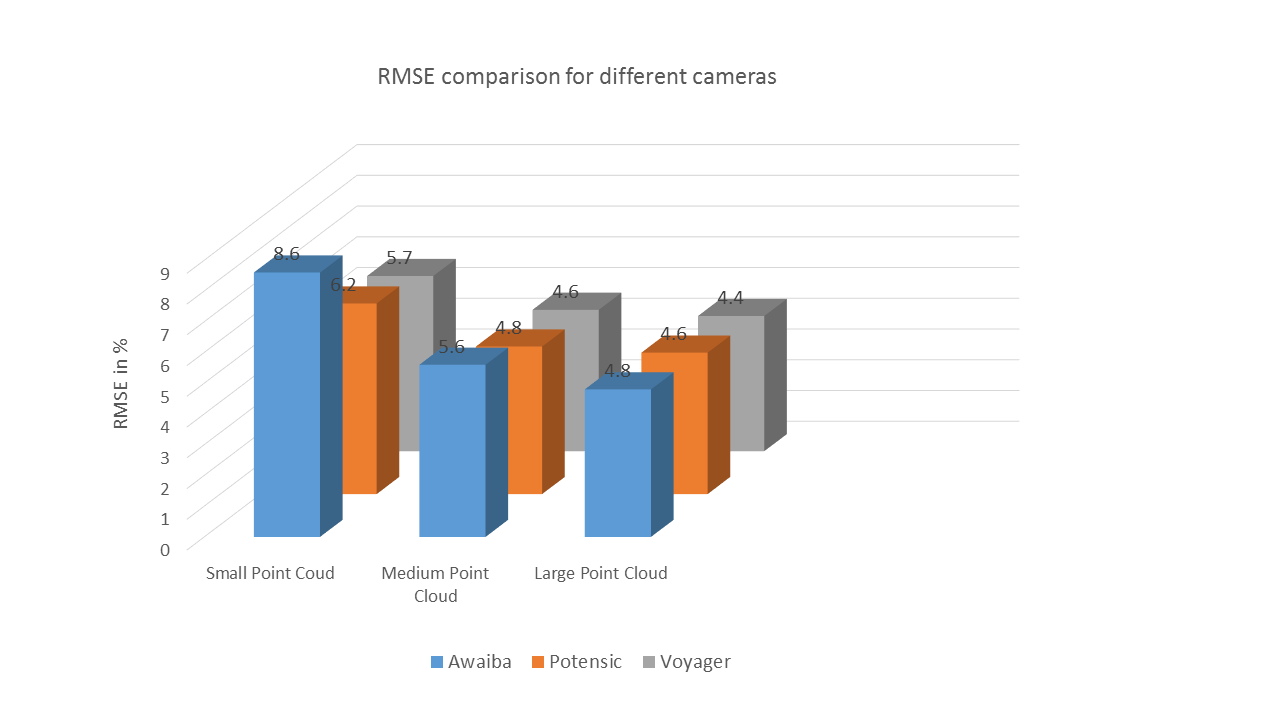}
\caption{ Percentage root mean square errors for different sized point clouds.}
\label{fig:15}       % Give a unique label
\end{figure*}
%Figure 16
\begin{figure*}
  \includegraphics[width=1\textwidth]{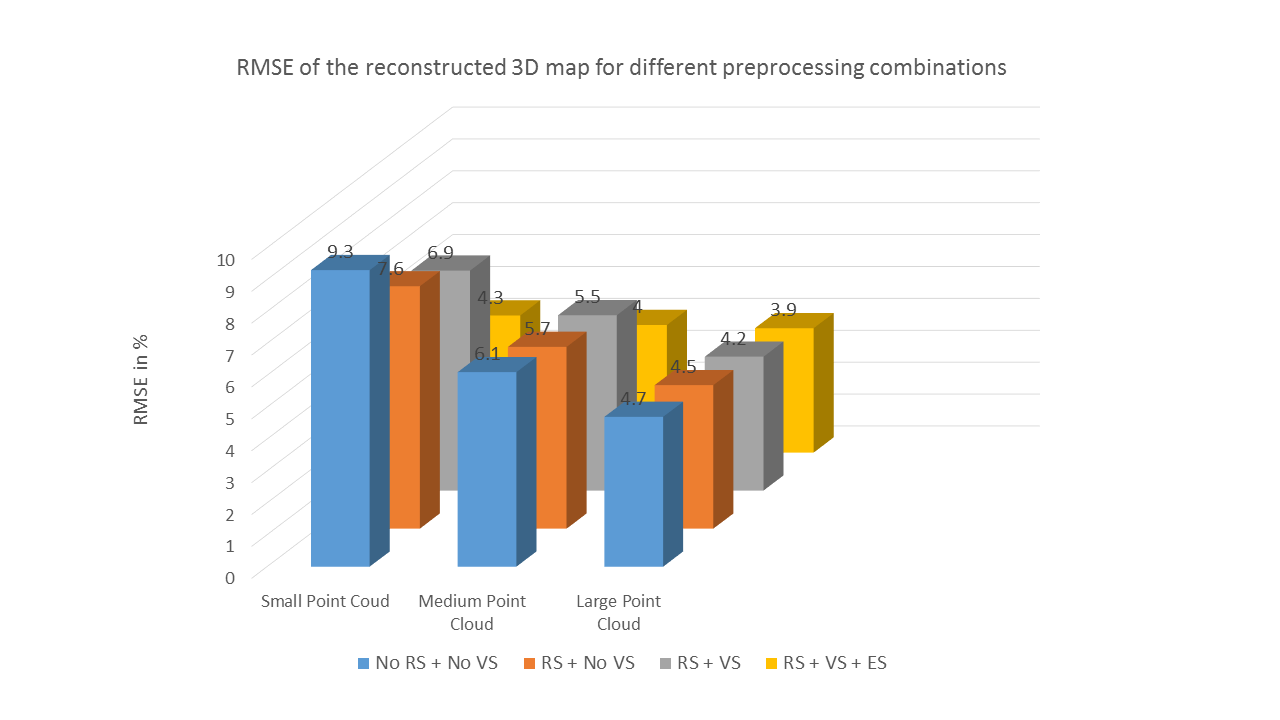}
\caption{ Effect of reflection suppression, vignetting correction and unsharp masking for three camera cases.}
\label{fig:16}       % Give a unique label
\end{figure*}
%Changed fig 14 to fig 10
Fig. 10 shows the effect of reflection distorion, vignetting suppression and edge sharpening on the final 3D map reconstruction for each of the three camera types.  We again observe that the NanEye camera generally performs poorly due to strong reflection, vignetting artefacts and image blur.  Images from Voyager and Potensic are less affected by reflection, blur and vignetting, and as such, reflection removal, edge sharpening and vignetting suppression operations have less influence on the accuracy of the final map. Frame-stitching is particularly affected by reflection, blur and vignetting due to failure of the feature matching process in images dominated by artefacts (see Fig. 10).  These artefacts also negatively affect shape-from-shading; the highlights of reflection artefacts cause artificial peak values in the depth map, and the distorting effects of vignetting affects the illumination and surface properties with which shape-from-shading estimates depth information.

\section{CONCLUSION}
Our proposed 3D map reconstruction framework demonstrates high precision mapping of a stomach simulator model, which could be used in more accurate 3D reconstructed surfaces for robotic capsule endoscope navigation and more improved diagnosis and treatment. As shown by the RMS error graphs, the framework works in all cases with less than 10$\%$ error.  The error rate increases with the number of frames in the point cloud, primarily due to the accumulated error of frame stitching.  In addition, reflection distortions, motion blur and vignetting artefacts reduce the performance of the 3D map reconstruction process.  Even with these problems, however, the RMS error is within an acceptable range for robotic capsule navigation or therapeutic purposes.\\ \\
In future work, we plan to estimate absolute depth with a stereo camera and combine this reading with shape from shading and image stitching to further improve the accuracy of our reconstructed 3D map.  Such a map of the GI tract might have dramatic implications for disease diagnosis, treatment, and other applications of active capsule endoscopes.

\section{Acknowledgements}
The authors thank to Abdullah Abdullah from RWTH Aachen University for his critical reading of the manuscript.

\section*{References}

\bibliography{references}

\end{document}